\documentclass[11pt]{article}

\usepackage[preprint]{acl}

\usepackage{times}
\usepackage{latexsym}

\usepackage[table]{xcolor}

\usepackage[T1]{fontenc}
\usepackage[utf8]{inputenc}

\usepackage{microtype}

\usepackage{inconsolata}

\usepackage{graphicx}

\usepackage{algorithm}
\usepackage{algorithmic}
\usepackage{amsmath}
\usepackage{listings}
\usepackage{xcolor}

\definecolor{codegreen}{rgb}{0,0.6,0}
\definecolor{codegray}{rgb}{0.5,0.5,0.5}
\definecolor{codepurple}{rgb}{0.58,0,0.82}
\definecolor{backcolour}{rgb}{0.95,0.95,0.92}

\lstdefinestyle{mystyle}{
    backgroundcolor=\color{backcolour},
    commentstyle=\color{codegreen},
    keywordstyle=\color{magenta},
    numberstyle=\tiny\color{codegray},
    stringstyle=\color{codepurple},
    basicstyle=\ttfamily\small,
    breakatwhitespace=false,
    breaklines=true,
    captionpos=b,
    keepspaces=true,
    numbers=left,
    numbersep=5pt,
    showspaces=false,
    showstringspaces=false,
    showtabs=false,
    tabsize=2
}

\usepackage{xspace}
\newcommand{\ourMethod}{\textbf{SCALE}\xspace}

\title{Do We Always Need Query-Level Workflows?
\\Rethinking Agentic Workflow Generation for Multi-Agent Systems}

\author{
Zixu Wang$^{\clubsuit \heartsuit}$,
Bingbing Xu$^{\clubsuit\,}$\thanks{Corresponding author.},
Yige Yuan$^{\clubsuit \heartsuit}$,
Huawei Shen$^{\clubsuit }$,
Xueqi Cheng$^{\clubsuit }$ \\
$^{\clubsuit}$ State Key Laboratory of AI Safety, Institute of Computing Technology, Chinese Academy of Sciences \\
$^{\heartsuit}$ University of Chinese Academy of Sciences \\
\texttt{\{wangzixu22s,xubingbing,yuanyige20z,shenhuawei,cxq\}@ict.ac.cn}
}

\begin{document}
\maketitle
\begin{abstract}
Multi-Agent Systems (MAS) built on large language models typically solve complex tasks by coordinating multiple agents through workflows. 
Existing approaches generates workflows either at task level or query level, but their relative costs and benefits remain unclear.
After rethinking and empirical analyses, we show that query-level workflow generation is not always necessary, since a small set of top-K best task-level workflows together already covers equivalent or even more queries. 
We further find that exhaustive execution-based task-level evaluation is both extremely token-costly and frequently unreliable.
Inspired by the idea of self-evolution and generative reward modeling, we propose a low-cost task-level generation framework \ourMethod, which means \underline{\textbf{S}}elf prediction of the optimizer with few shot \underline{\textbf{CAL}}ibration for \underline{\textbf{E}}valuation instead of full validation execution. 
Extensive experiments demonstrate that \ourMethod maintains competitive performance, with an average degradation of just 0.61\% compared to existing approach across multiple datasets, while cutting overall token usage by up to 83\%.

\end{abstract}

% Section 1
\section{Introduction}

Large Language Model (LLM)-based multi-agent systems (MAS) have recently emerged
as a powerful paradigm for solving complex reasoning, coding, and decision-making tasks \citep{zhang2024aflow,zhang2024agent_prune,zhuge2024gptswarm,niu2025flow}.
By decomposing a task into multiple interacting agents and organizing their
collaboration through agentic workflows, MAS can substantially extend the capabilities of a single agent.

\begin{figure}
    \centering
    \includegraphics[width=\linewidth]{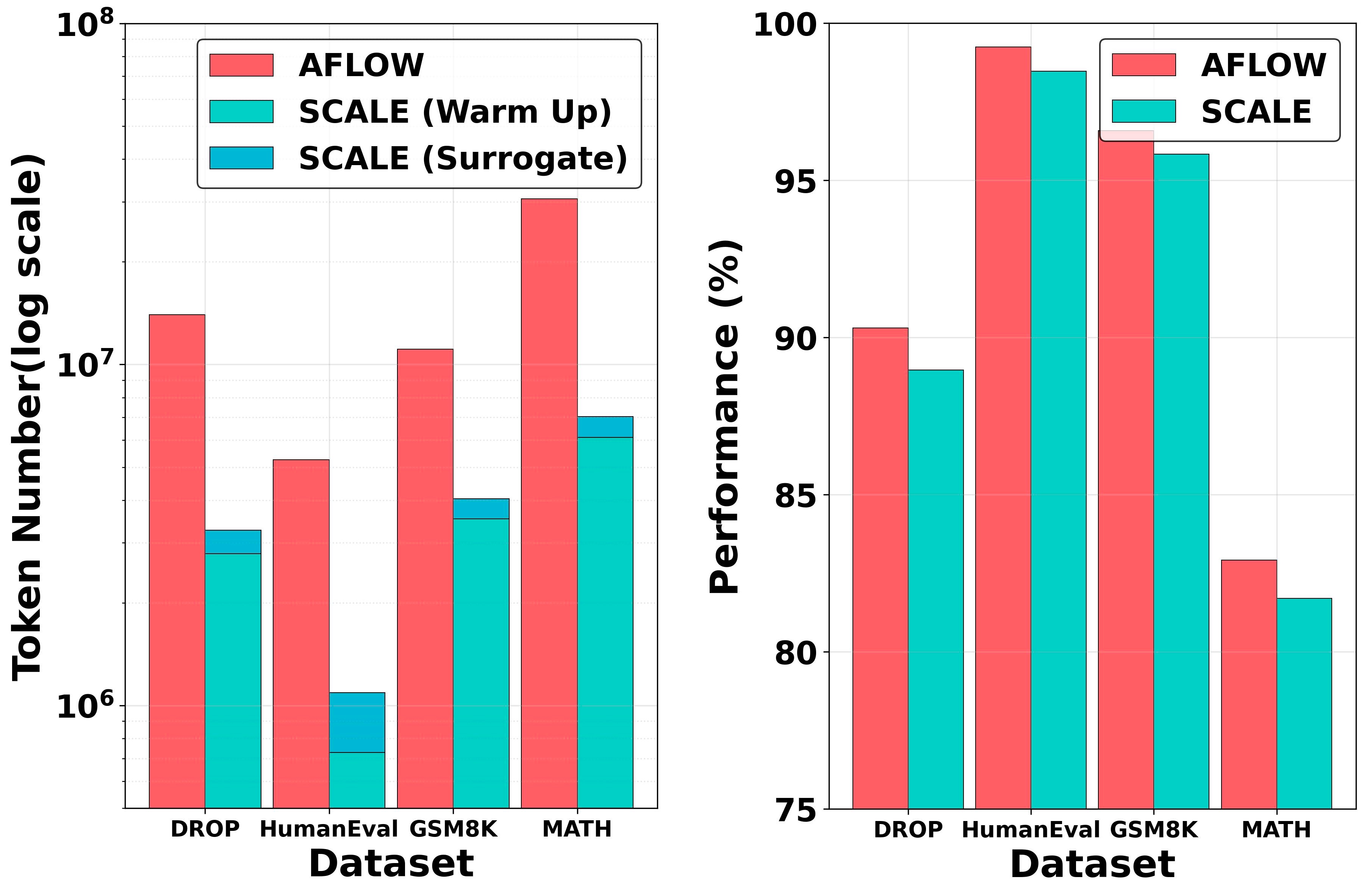}
    \caption{Comparison of Aflow and our rethought task-level workflow generation framework. Left: total token number during workflow generation (log-scale axis). Right: final test performance. Our method \ourMethod achieves comparable performance while significantly reducing token number.}
    \label{fig:cost_acc_comparision}
    \vspace{-0.5em}
\end{figure}

Based on the granularity of workflow construction, agentic workflow generation methods fall into two categories: task-level and query-level approaches.
Task-level approaches, such as such as search-based Aflow\cite{zhang2024aflow} and learning-based GPTSwarm\cite{zhuge2024gptswarm} and AgentPrune\cite{zhang2024agent_prune}, generate a single workflow intended to perform well across an entire dataset or task distribution.
% Despite differing implementations, they share a loop paradigm: iteratively generating workflows, evaluating them on the full validation set, and using feedback to guide subsequent generation. 
% Once identified, the best workflow is reused for all test queries. 
However, this generality comes at a high cost: evaluation dominates computation, as each candidate requires full execution over the validation set. 
As shown in Figure~\ref{fig:cost_acc_comparision}, the whole Aflow's generation process on four benchmarks consumes approximately $10^{6}\!-\!10^{8}$ LLM tokens.

In parallel, query-level approaches generate a separate workflow for each input query\citep{ye2025mas_gpt,wang2025scoreflow,gao2025flowreasoner}. 
For every query, the system constructs a customized multi-agent workflow, allowing the agent roles and interaction patterns to adapt to the specific problem. 
This design aims to better handle heterogeneous queries and can yield strong per-query performance.
However, this adaptivity comes with clear costs. 
A new workflow must be generated for every query, which introduces substantial inference overhead. 
For many simple or similar inputs, such query-level generation may be unnecessary, providing limited gains relative to its train time computational cost and test time generation cost.

From the characteristics of these two paradigms, we raise  two fundamental questions that have not been systematically examined as shown in Figure~\ref{fig:process_rethinkig}.
First, \emph{Is query-level workflow generation always necessary?}
Second, \emph{Is high-cost evaluation in  task-level workflow generation necessary?}
For the first question, we show that a small set of top-k task-level workflows already achieves strong query coverage comparing to query-level method's performance. It indicates that query-level workflow generation is not always necessary in practice. 
For the second, we find that exhaustive execution-based evaluation of task-level workflows is both extremely expensive and frequently unreliable.

Inspired by the idea of self-evolution and generative reward modeling, we propose a low-cost task-level generation framework \ourMethod, which means \underline{\textbf{S}}elf prediction of the optimizer with few shot \underline{\textbf{CAL}}ibration for \underline{\textbf{E}}valuation instead of full validation execution. 
By leveraging the inherent evaluative ability of LLM-based optimizers, \ourMethod makes self predictions in a generative manner and calibrates them using few shot executions, thereby achieving highly reliable predictions with minimal token cost. 
Experimental analysis further shows that the calibrated self predictions in \ourMethod closely approximate true execution scores, it achieves a low MAE of 0.16 and maintain consistent ranking with a Pearson correlation of 0.52 (range: $[-1, 1]$), further validating reliability.
Overall, our contributions are threefold as shown below:
\begin{itemize}
\item \textbf{Rethinking Insights:} We present a new empirical rethinking of workflow generation in multi-agent systems.
Our analysis yields two main findings:
(1) Query-level methods is not always necessary in practice.
(2) Exhaustive execution-based evaluation in task-level approaches is both costly and unreliable.

\item \textbf{Improved Framework:}
Motivated by these observations and inspired by self-evolution, we develop a low-cost and effective framework \ourMethod  for task-level workflow generation.
Instead of exhaustively executing candidate workflows on the full validation set, our approach combines the LLM-based optimizer's self prediction with few shot calibration to evaluate workflows efficiently.

\item \textbf{Empirical Validation:}
Extensive experiments demonstrate that \ourMethod maintains competitive performance, with an average degradation of just 0.61\% compared to existing approach across multiple datasets, while cutting overall token usage by up to 83\%.
\end{itemize}

% Section 2
\section{Preliminaries}
\label{sec:prelim}

\subsection{Agentic MAS Workflow}
\label{sec:mas_workflow}
We consider a task $\mathcal{T}$ given as a dataset of queries $q \in \mathcal{D}$, and an agentic MAS workflow $W \in \mathcal{W}$ that orchestrates a LLM-based MAS to produce an answer $y$. 
Formally, we formalizes the workflows space as:
\begin{equation}
\begin{aligned}
\mathcal{W} = \bigl\{ (&P_1, \dots, P_n, E, O_{\theta_1}, \dots, O_{\theta_n}) \\ \;\mid\; & P_i \in \mathcal{P},\; E \in \mathcal{E},\; O_{\theta_i} \in \mathcal{O} \bigr\}
\end{aligned}
\label{eq:aflow_space}
\end{equation}
where $n$ is the number of agents, 
$\mathcal{P}$ is the prompt space, 
$\mathcal{E}$ is the information control flow that governs the execution order, data dependencies, and communication among these agents, which can be represented in various forms: as executable code (e.g.\citealp{hu2024adas, zhang2024aflow, xu2025robustflow}) or as directed acyclic graphs(e.g. \citealp{zhuge2024gptswarm, zhang2024agent_prune, wang2025agentdropout, zhang2025maas}). 
$\mathcal{O}$ is a set of predefined LLM-based agents parameterized by $O_{\theta_i}$. 
The prompt $P_i$ and parameters $\theta_i$ enable the agent to adapt its behaviors to the task at hand, such as \emph{Review}\cite{yao2022react}, \emph{Ensemble}\cite{liang2024debate}, or \emph{Self-Correction}\cite{shinn2023self_reflect}.

At a high level, a workflow is a series of agent calls to answer a certain query $y = W(q)$.
Given a task-specific evaluator $s(\cdot,\cdot)$ (e.g., exact match, pass@\!1), the performance of $W$ on a query $q$ is measured by $s(W(q), q) \in [0,1]$.

\subsection{Agentic MAS Workflow Generation}
\label{sec:agentic_optimization}

\begin{figure*}[htbp]
    \centering
    \includegraphics[width=\linewidth]{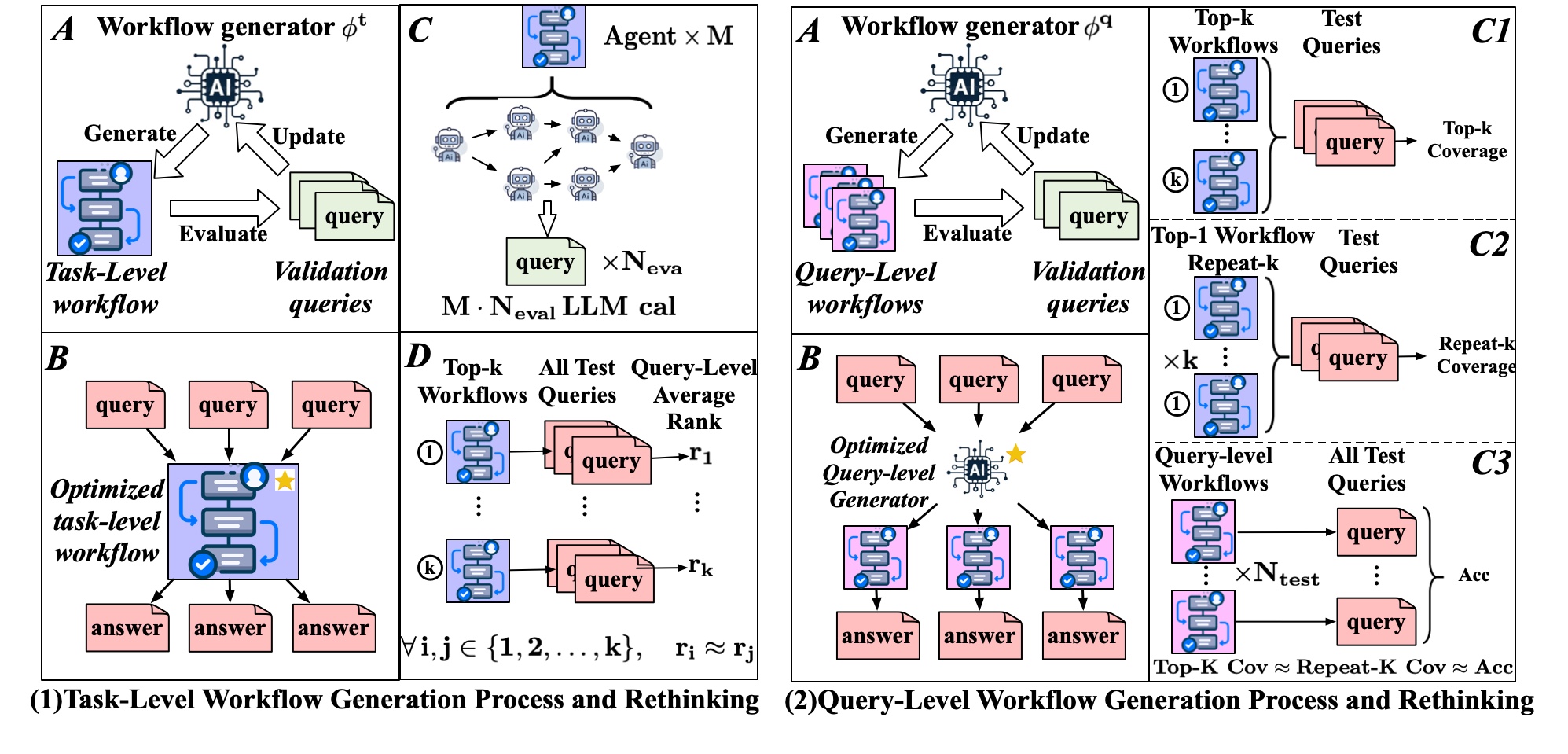}
    \vspace{-1em}
    \caption{Task-level vs. Query-level workflow generation on their process and rethinking.(1)Task-level generation.(1)A shows searching/training: the generator generates a single workflow using validation queries; (1)B shows inference: the optimized workflow is reused for all test queries. (1)C–D present our rethinking: (1)C shows that repeated full-set evaluations is very token-costly, and (1)D shows top-k workflows have very similar query-level ranks. (2)Query-level generation. (2)A shows training: a workflow is generated per query; (2)B shows inference: producing customized workflows for each input. (2)C1–C3 summarize our rethinking on query-level workflows: top-k task-level workflows, repeat-k runs of the top-1 workflow, and true query-level generation yield comparable coverage/performance.
    }
    \label{fig:process_rethinkig}
    \vspace{-1.5em}
\end{figure*}

A number of recent methods have been proposed for agentic workflow generation, which can be grouped into two paradigms according to the granularity at which workflows are generated: task-level approaches and query-level approaches.

\subsubsection{Task-level workflow generation}
\label{sec:task_level_prelim}

Task-level methods aim to generate a single workflow $W^\ast$ that performs well on a distribution of queries from the same task as shown in (1)A and (1)B of Figure~\ref{fig:process_rethinkig}. 
Given a validation set $\mathcal{D}_{\mathrm{val}} = \{q_i\}_{i=1}^N$ sampled from the task $\mathcal{T}$, 
the objective is:
\begin{equation}
\begin{aligned}
    W^\ast_{\mathrm{task}} =&\arg\max_{W \in \mathcal{W}}
\; S^{exec}(W, \mathcal{D}_{val})
\end{aligned}
\label{eq:task_level_obj}
\end{equation}
where $\mathcal{W}$ is the workflow search space,  and $S^{exec}(W, \mathcal{D}_{\mathrm{val}}) = \frac{1}{|\mathcal{D}_{\mathrm{val}}|}
\sum_{q \in \mathcal{D}_{\mathrm{val}}}
s(W(q), q)$ denotes the average execution score of workflow $W$ on dataset $\mathcal{D_{\mathrm{val}}}$.

Despite implementation differences, these methods share the same closed-loop paradigm. First, generating candidate workflows through a task-level workflow generator $\phi^t$. Second, evaluating the generated workflow on $\mathcal{D}_{\mathrm{val}}$. Finally, updating the model $\phi^t$ using evaluation as a feedback.
Aflow~\cite{zhang2024aflow} uses a LLM as optimizer $\phi^t$ and improve the initial workflow through an MCTS-style loop. 
AgentPrune~\cite{zhang2024agent_prune} use a graph model as workflow generator $\phi^t$ and learn it through reinforcement learning methods to generate better workflows.

Despite good performance, their evaluation is extremely costly, since each candidate's evaluation requires the MAS workflow to execute on the full validation set. 
The token number scales with the validation dataset size and agent numbers resulting in a sharp increase as the loop continues.

\subsubsection{Query-level workflow generation}
\label{sec:query_level_prelim}

As shown in (2)A and (2)B of Figure~\ref{fig:process_rethinkig}, query-level methods instead learn a query-level workflow generator $\phi^q$ that maps each query $q$ to its own workflow $W_q= \phi^q(q)$
and the optimization objective is to maximize expected performance over the
validation set:
\begin{equation}
\phi^{q,\ast}
=
\arg\max_{\phi^q}
\;
\frac{1}{|\mathcal{D}_{\mathrm{val}}|}\sum_{q \in \mathcal{D}_{\mathrm{val}}}
\big[
s(W_q(q),\; q)
\big]
\label{eq:query_level_obj}
\end{equation}
% where $W_q(q)$ denotes executing the workflow produced by $\phi^q(q)$ for query $q$.

Existing approaches differ mainly in how $\phi^q$ is trained.  
MAS-GPT~\cite{ye2025mas_gpt} adopts supervised fine-tuning on a curated dataset of query–workflow pairs.
ScoreFlow~\cite{wang2025scoreflow} optimizes the workflow generator $\phi^q$ using a preference-based optimization approach which enhances the original DPO~\cite{rafailov2023dpo}. 
For many simple or structurally similar inputs, such query-level workflow generation may be unnecessary, providing limited gains relative to its test-time generation cost.

\section{Rethinking Agentic Workflow Generation for Multi-Agent Systems}
\label{sec:rethinking}

Our rethinking is twofold.
First, as shown in (2)C1-C3 of Figure~\ref{fig:process_rethinkig},  we rethink the necessity of query-level workflow generation. 
Second,  as shown in (1)C and (1)D of Figure~\ref{fig:process_rethinkig}, we rethink task-level methods, arguing that execution on validation set for evaluation is both token-costly and unreliable.

\begin{table}[htbp]
\centering
\small
\setlength{\tabcolsep}{2.5pt}
\begin{tabular}{l c c cc cc}
\hline
& \multicolumn{5}{c}{\textbf{Aflow}} & \textbf{S.Flow} \\
\textbf{Dataset} 
& \multicolumn{1}{c}{\textbf{Top-1}}
& \multicolumn{2}{c}{\textbf{Top-5}}
& \multicolumn{2}{c}{\textbf{Repeat-5}} 
& \textbf{\textbf{All}} \\
& 
\textbf{Perf}
& \textbf{Perf} & \textbf{Cov} 
& \textbf{Perf} & \textbf{Cov} & \textbf{Perf} \\
\hline
\textbf{DROP} & 90.30 &  89.81  &  93.87    &  90.04  & 92.45 &   91.48  \\
\textbf{HumanEval} & 99.24 & 98.17 & 100.00 & 97.82 & 99.24 & 98.91 \\
\textbf{GSM8K} & 96.58 & 95.89 & 97.35 & 96.17 & 96.87 &   97.79   \\
\textbf{MATH} & 82.92 & 79.84 & 87.04 & 82.81 & 83.39 &  84.35 \\
\hline
\end{tabular}
\caption{
    Comparison of task-level and query-level workflow effectiveness.
    \textbf{Perf} denotes average test performance (\%).
    \textbf{Cov} denotes coverage (\%) of test queries.
    \textbf{S.Flow} denotes the query-level method ScoreFlow.
}
\label{tab:task-coverage}
\vspace{-2em}
\end{table}
\subsection{Is Query-level Workflow Generation Always Necessary?}
\label{sec:topk_coverage}

Query-level methods offers fine-grained adaptivity, but it introduces considerable test-time generation cost that task-level methods don't have. This raises a fundamental question:

\emph{Is query-level workflow generation always necessary to achieve strong performance?}

For each dataset, we evaluate the following settings.(1)Task-level Top-1: We report the test performance of  Aflow's~\cite{zhang2024aflow} best generated workflow.
(2)Task-level Top-5: We report average test performance and coverage of Aflow's top-5 workflows. \emph{Coverage} is defined as the fraction of test queries that are covered by these workflows.
A query is counted as covered if at least one of these workflows answers it correctly. 
(3)Repeat-5 of Top-1:
We execute the Aflow's top-1 workflow on test set 5 times. We report the average performance and coverage on test queries.
This isolates the effect of the test-time execution stochasticity.
(4)Query-level workflows:
A dedicated workflow is generated for each query using a representative query-level method ScoreFlow~\cite{wang2025scoreflow}, and we report its average test performance.

As shown in Table \ref{tab:task-coverage}, we obtain three main observations.
First, a single Top-1 task-level workflow already performs strongly, indicating substantial structural sharing across queries.
Second, Top-5 gains a clear increase in query coverage even more than query-level method's performance, meaning that improvements can come from gathering few candidate task-level workflows rather than generating single strictly better workflows.
Third, Repeat-5 achieves coverage comparable to query-level method, showing that much of the gain by query-level method compared to task-level can by covered by stochastic execution rather than diverse workflow structures.
\begin{figure}[htbp]
	\centering
	\includegraphics[width=\linewidth]{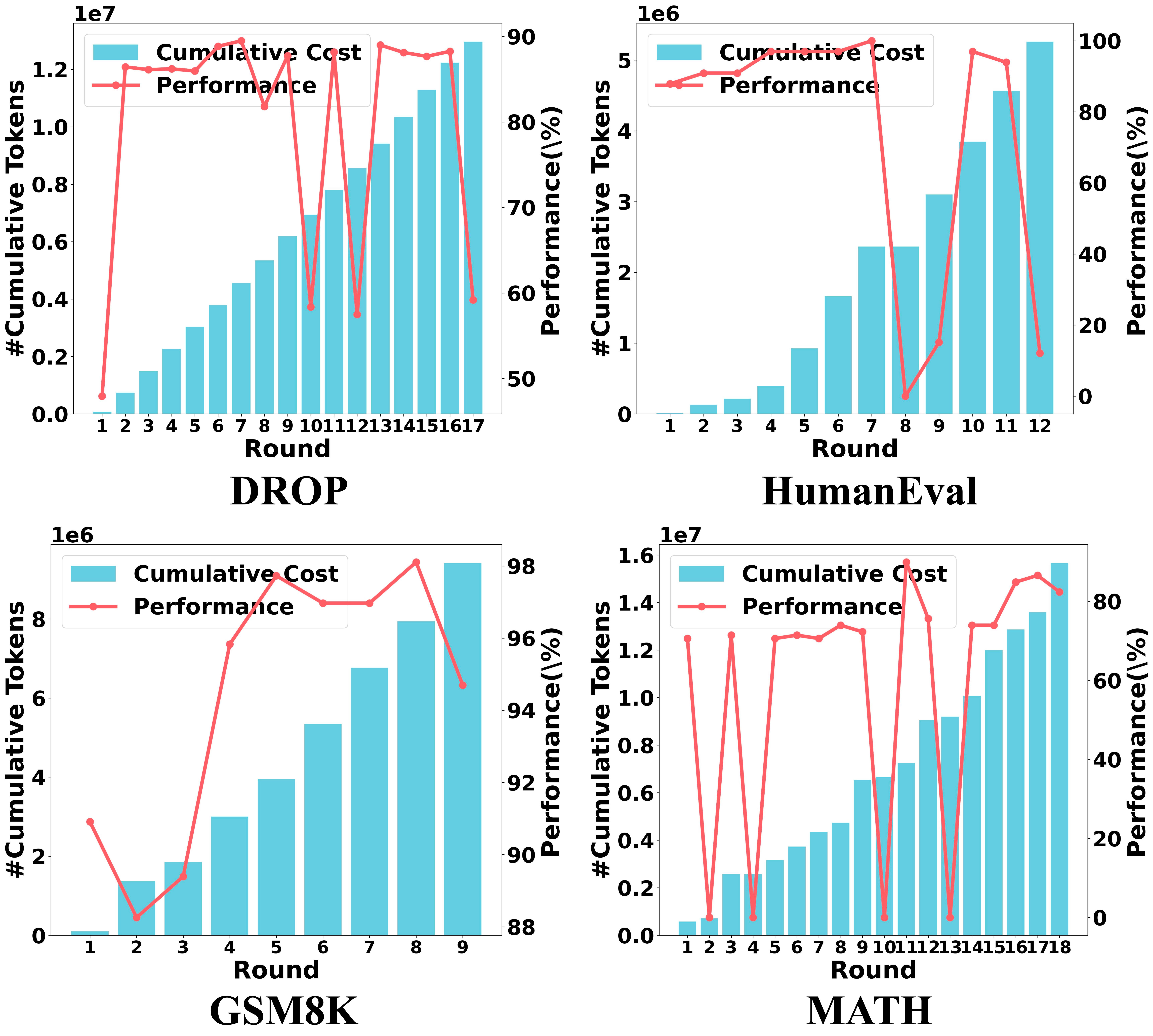}
    \caption{Cumulative Token Number v.s. Performance during Aflow's task-level workflow generation process.}
    \label{fig:token_cost}
    \vspace{-1.5em}
\end{figure}

Taken together, these results suggest that most benefits attributed to query-level method can already be achieved by a small pool of reusable task-level workflows or even repeated execution of a single strong task-level workflow. 
The main advantage can come from coverage and stochasticity, not necessarily in query-level workflows.

\subsection{Is High-Cost Evaluation in  Task-Level Workflow Generation Necessary?}
\label{sec:token_cost}

In this subsection, we revisit the evaluation in task-level workflow generation from two complementary perspectives:
(1) how many tokens actually incurs in exhaustive execution-based evaluation, and (2) whether such costly
evaluation truly leads to meaningfully different workflows.

\subsubsection{How many evaluation tokens are incurred during task-level workflow generation?}
% 每张子图包含两条折线和一个bar； 横轴是搜索过程中的不同round，折现分别表示valid-test-set的acc; bar表示目前累计的搜索总cost；

% While the performance differences between top-ranked workflows are small, the cost required to identify them is substantial.
We analyze the cumulative evaluation token number incurred during Aflow's workflow generation.
Figure~\ref{fig:token_cost} illustrates the relationship between evaluation token number and performance across each generated candidate workflow.
For each benchmark, we plot the test performance achieved by candidate workflows, together with the cumulative token number which is defined as the total tokens used to evaluate all candidate workflows generated up to and including the current round.

Figure~~\ref{fig:token_cost} shows that cumulative evaluation token number grows fast continuously with search rounds, while test performance quickly saturates and yields only marginal or negative gains afterward.
This mismatch is evident: the cost is exploding but the corresponding performance improvement is minimal or even negative. 
These results indicate that the prevailing task-level evaluation paradigm is both expensive and unreliable in the high-performance regime, motivating the need for cheaper and more reliable task-level workflow evaluation.

\subsubsection{Do high-cost task-level evaluations actually distinguish better workflows?}
% \begin{table}[htbp]
% \centering
% \small
% \setlength{\tabcolsep}{4pt}
% \begin{tabular}{lcccc}
% \hline
% \textbf{Benchmark} & 
% \textbf{Workflow} & 
% \textbf{Acc (\%)} & \textbf{CR} & \textbf{DR} \\
% \hline
% \textbf{DROP} & Top-1 & 90.30 & 1.13 & 1.07 \\
%      & Top-2 & 90.23 & 1.13 & 1.07 \\
%      & Top-3 & 90.02 & 1.14 & 1.07 \\
%      & Top-4 & 89.84 & 1.15 & 1.08 \\
%      & Top-5 & 88.69 & 1.25 & 1.10 \\
% \hline
% \textbf{HumanEval} & Top-1 & 99.24 & 1.00 & 1.00 \\
%           & Top-2 & 98.47 & 1.04 & 1.01 \\
%           & Top-3 & 97.78 & 1.05 & 1.02 \\
%           & Top-4 & 97.69 & 1.05 & 1.02 \\
%           & Top-5 & 97.66 & 1.07 & 1.03 \\
% \hline
% \textbf{GSM8K} &Top-1 & 96.58 & 1.02 & 1.01 \\
%       & Top-2 & 96.30 & 1.03 & 1.01 \\
%       & Top-3 & 96.02 & 1.03 & 1.01 \\
%       & Top-4 & 95.83 & 1.04 & 1.02 \\
%       & Top-5 & 94.69 & 1.09 & 1.03 \\
% \hline
% \textbf{MATH} & Top-1 & 82.92 & 1.07 & 1.04 \\
%      & Top-2 & 82.51 & 1.08 & 1.04 \\
%      & Top-3 & 81.48 & 1.12 & 1.06 \\
%      & Top-4 & 81.07 & 1.13 & 1.06 \\
%      & Top-5 & 71.19 & 1.55 & 1.16 \\
% \hline
% \end{tabular}
% \caption{
% Performance and ranking statistics of the top-$5$ task-level workflows generated by Aflow across four benchmarks.
% \textbf{Acc} denotes average test accuracy.
% \textbf{CR} and \textbf{DR} denote average competition rank and dense rank,
% respectively, computed over test queries.
% }
% \label{tab:top5_workflows}
% \end{table}
\begin{figure}
    \centering
    \includegraphics[width=\linewidth]{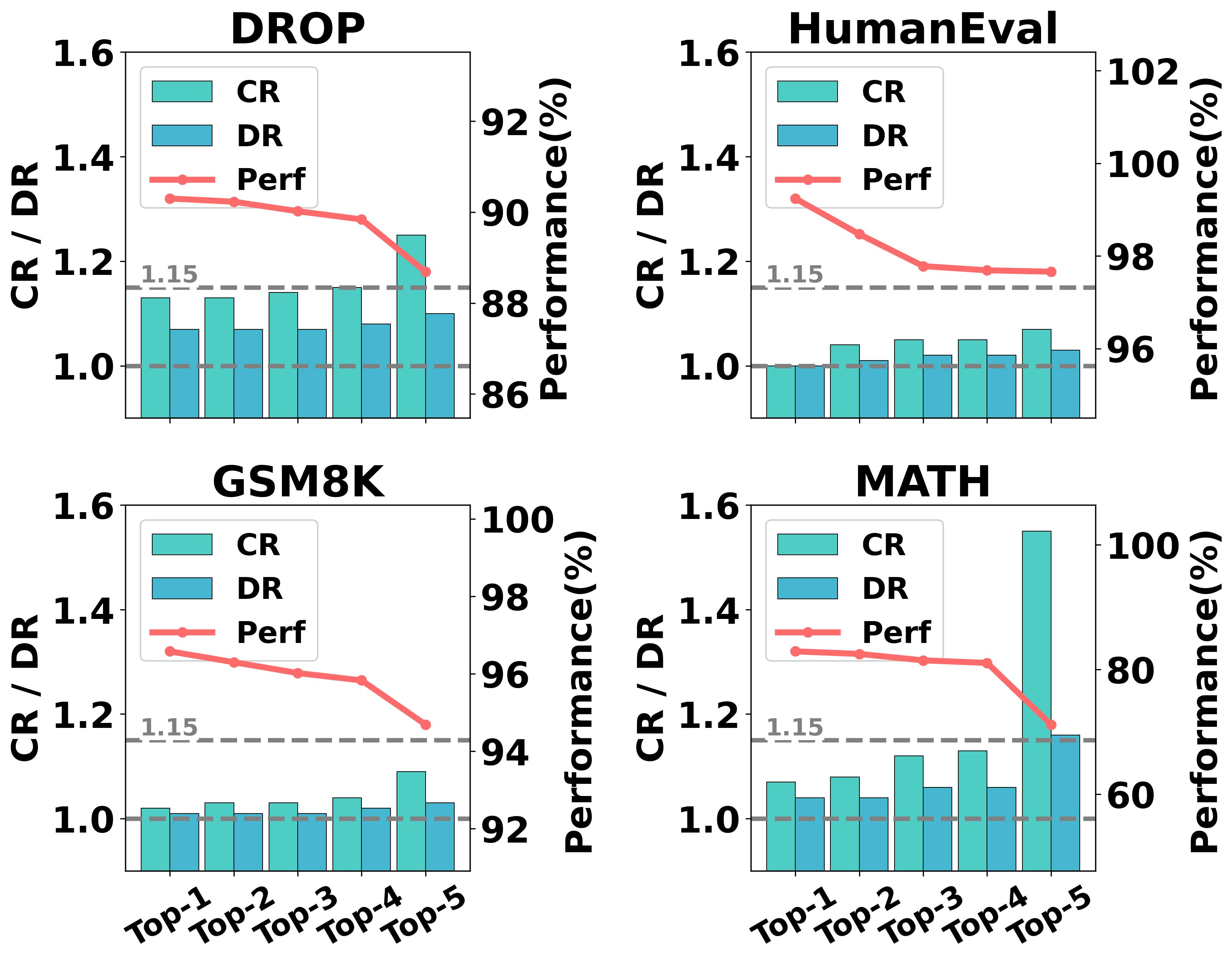}
    \caption{
    Performance and ranking statistics of the top-$5$ task-level workflows generated by Aflow across four benchmarks.
    \textbf{Perf} denotes average test performance.
    \textbf{CR} and \textbf{DR} denote average competition rank and dense rank,
    respectively, computed over test queries.
    }
    \label{fig:top_5_workflows}
    \vspace{-1em}
\end{figure}
To further examine the efficacy of Aflow's costly evaluation process, we analyze the Top-5 workflows, defined as the five candidates with the highest validation performance across all candidate workflows produced in one complete run.

Beyond test performance, we also analyze query-level ranking. Concretely, all Top-5 workflows are executed and ranked for each test query. 
Then Top-5 workflows' competition rank (\textbf{CR}) which allows ties and dense rank (\textbf{DR}) are averaged across queries. 
This reveals how consistently one workflow beats another. 
If the evaluation are strongly discriminative, these ranks would show clear separation among Top-5 workflows.

Figure~\ref{fig:top_5_workflows} shows that the Top-5 (especially Top-4) task-level workflows obtained by Aflow exhibit very similar performance. Their performances vary only slightly across all benchmarks, while both CR and DR remain close to 1 with minimal variation, which indicates that expensive full validation provides limited benefit in identifying substantially better workflows.

\subsection{Rethinking Results}

In Section~\ref{sec:topk_coverage}, we observed that query-level workflow generation is not always necessary, because a small set of task-level workflows or even repeated executions of a single workflow already covers more queries. 
In Section~\ref{sec:token_cost}, we further showed that exhaustive execution-based task-level evaluation is extremely costly while providing limited discriminative benefit.

Together, these findings show that current agentic workflow methods waste a lot of computation either generating unnecessary query-level workflows or evaluating task-level workflows that have almost the same performances. This motivates a new framework for task-level workflow generation that avoids both query-level generation and high-cost full-execution-based evaluation.

% Section 3 (Core Analysis)

\section{Methodology}
\subsection{Motivation}
% Our rethinking leads to a clear objective: enable task-level workflow generation that is both low-cost and reliable.
We propose a low-cost task-level generation framework \ourMethod, which means \underline{\textbf{S}}elf prediction with few shot \underline{\textbf{CAL}}ibration  for \underline{\textbf{E}}valuation.
Inspired by the self-evolution paradigm and generative reward modeling in agentic systems, we treat the workflow optimizer as a self-predictor. In other words,  the same LLM that generates workflows is also prompted to estimate the candidate workflow's expected performance.

To reduce overconfidence, we separate generation and evaluation prompts and obtain scores in a dedicated evaluation context. 
We calibrate self predictions using execution results from few shot queries, typically 1–3\% of the validation set.
As a result, \ourMethod enables task-level evaluation without full validation execution, while maintaining reliable workflow scoring and ranking.

\subsection{Overall framework}
Our method \ourMethod operates in two stages:
a short warm-up stage with full execution for evaluation, followed by a surrogate evaluation stage.
Specifically, we use Aflow~\cite{zhang2024aflow} for a few steps as the warm-up stage, and then instead of repeatedly computing
$S^{\mathrm{exec}}(W,\mathcal{D}_{\mathrm{val}})$, we estimate workflow quality using a self prediction score calibrated by few shot execution as the surrogate evaluation stage.

\subsubsection{Warm-up Stage}
\label{sec:warmup}
We begin with $M$ warm-up rounds. 
Concretely, starting from a single question-answering LLM agent call as the initial workflow $W_1$ with execution score $S^{\mathrm{exec}}_1$, we run an MCTS-style loop for $t=\{1,\dots,M\}$ consisting of four steps:

\textbf{1. Selection.}
Given the existing workflows and scores $\{S^{\mathrm{exec}}_i\}_{i=1}^t$, we select a parent workflow using a soft mixed policy:
\begin{equation}
P_i= \lambda \cdot \frac{1}{t} + (1 - \lambda) \cdot 
\frac{ \exp\big(\alpha (S^{\mathrm{exec}}_i - S^{\mathrm{exec}}_{\max})\big) }{ 
\sum_{j=1}^t \exp\big(\alpha (S^{\mathrm{exec}}_j - S^{\mathrm{exec}}_{\max})\big) },
\label{eq:aflow_select}
\end{equation}
where $S^{\mathrm{exec}}_{\max}=\max_j S^{\mathrm{exec}}_j$, and $\lambda,\alpha$ control the exploration–exploitation trade-off.

\textbf{2. Expansion.}
A new workflow is generated by editing $W_i$ using the LLM-based optimizer:
\begin{equation}
W_{t+1}=\phi\!\left(W_i;\, P^{\mathrm{optimizer}}_{t+1}\right)
\label{eq:aflow_expand}
\end{equation}
The optimizer prompt $P^{\mathrm{optimizer}}_{t+1}$ is dynamically built from local experience $E^{\mathrm{local}}_i$ and global experience $E^{\mathrm{global}}$ introduced in backpropagation step.

\textbf{3. Evaluation.}
We execute $W_{t+1}$ on the validation set 
$S^{\mathrm{exec}}_{t+1}= \frac{1}{|\mathcal{D}_{\mathrm{val}}|}\sum_{q \in \mathcal{D}_{\mathrm{val}}}s\!\left(W_{t+1}(q), q\right)$.

\textbf{4. Backpropagation}
The evaluation result updates both local and global experience. 
For $W_i$ the local experience is
$E^{\mathrm{local}}_i \leftarrow E^{\mathrm{local}}_i \cup \!\left(e_i^{t+1}\right)$
where $e_i^{t+1}=\Big((W_i, S_i),\, \Delta_{i}^{t+1},\, (W_{t+1}, S_{t+1})\Big)$ and $\Delta_{i}^{t+1}$ denotes the optimizer's natural language description of the edit from $W_i$ to $W_{t+1}$. We also maintain a global experience:
$E^{\mathrm{global}} \leftarrow E^{\mathrm{global}} \cup \{(W_{t+1}, S^{\mathrm{exec}}_{t+1})\}$.
These experience is reused to update future optimizer prompts and the selection policy. 

\subsubsection{Surrogate Evaluation Stage}
\label{sec:lowcost}

After warm-up stage, we continue the loop but the subsequent workflows are evaluated through self prediction with few shot execution calibration.
At iteration $t>M $, we select and expand using Equation~\ref{eq:aflow_select}
and Equation~\ref{eq:aflow_expand} to get the newly generated workflow $W_{t+1}$. We evaluate it with our method: 

\paragraph{Self prediction}
Firstly, we query the optimizer itself under a dynamically dedicated evaluation prompt $P^{\mathrm{optimizer}}_{t+1}$ to obtain the prediction:
\begin{equation}
\label{eq:selfpred_final}
S^{\mathrm{pred}}_{t+1}=S^{\mathrm{pred}}(W_{t+1}) = \phi\!\left(W_{t+1} ;\ P^{\mathrm{eval}}_{t+1}\right)
\end{equation}
where $\phi$ is the same LLM-based optimizer used for workflow expansion in Equation~\ref{eq:aflow_expand}.
The evaluation prompt $P^{\mathrm{eval}}_{t+1}$ is separated from the optimization prompt $P^{\mathrm{optimizer}}_{t+1}$ to reduce overconfidence and prompt entanglement. The full template of $P^{\mathrm{eval}}$ is provided in Appendix~\ref{app:prompt_template}.

\paragraph{Few shot execution calibration}
To reduce bias, we execute $W_{t+1}$ only on a small subset
$\mathcal{D}_{\mathrm{few}}\subset\mathcal{D}_{\mathrm{val}}$ with $|\mathcal{D}_{\mathrm{few}}|\ll|\mathcal{D}_{\mathrm{val}}|$:
\begin{equation}
\label{eq:fewshot_final}
\begin{aligned}
S^{\mathrm{few}}_{t+1}
= \frac{1}{|\mathcal{D}_{\mathrm{few}}|} \sum_{q\in \mathcal{D}_{\mathrm{few}}} s\!\left(W_{t+1}(q), q\right)
\end{aligned}
\end{equation}
The sampling of $\mathcal{D}_{\text{few}}$ is guided by the full-execution statistics collected during warm-up. 
With  warming up workflows $\{W_m\}_{t=1}^{M}$, for each validation query $q \in \mathcal{D}_{\mathrm{val}}$, we compute its empirical warm-up score 
$\bar{s}(q)= \frac{1}{M} \sum_{t=1}^{M} s\!\left(W_m(q), q\right)$ 
which reflects how well warm-up workflows already solve $q$.
We then partition the range of $\bar{s}(q)$ into $K$ bins $\{B_k\}_{k=1}^K$ (e.g., $K=10$), where
$B_k = \bigl\{ q \in \mathcal{D}_{\mathrm{val}} \,\big|\, \bar{s}(q) \in I_k \bigr\}$ 
and $\{I_k\}_{k=1}^K$ are disjoint score intervals covering $[0,1]$.
Let $n_k = |B_k|$ be the number of queries in bin $k$.
We define a bin-level sampling distribution by a softmax over bin counts 
$p_k = \frac{\exp\!\big(\gamma n_k\big)} {\sum_{j=1}^{K} \exp\!\big(\gamma n_j\big)}, k=1,\dots,K$
, where $\gamma > 0$ is the sampling temperature.

Given a target budget $|\mathcal{D}_{\mathrm{few}}| = \rho\,|\mathcal{D}_{\mathrm{val}}|$ with $\rho \in [0.01, 0.03]$, we sample queries without replacement by first sampling a bin index $k$ from the categorical distribution $\{p_k\}$,
and then sampling a query $q$ uniformly from $B_k$.
Repeating this procedure until $|\mathcal{D}_{\mathrm{few}}|$ is reached yields a few shot subset that preserves the warm-up difficulty distribution while covering both easy and hard queries. 
% in a statistically representative way.

\paragraph{Calibrated surrogate score}
The final score used to evaluate the workflow is:
\begin{equation}
\label{eq:calibratedscore_final}
\begin{aligned}
    \widehat{S}_{t+1}
    &=(1-\alpha_{t+1})\, S^{\mathrm{pred}}_{t+1}
+ \alpha_{t+1} \, S^{\mathrm{few}}_{t+1} 
\end{aligned}
\end{equation}
where $\alpha_{t+1}\in[0,1]$ controls how strongly we trust the few shot
execution relative to self prediction.

In practice, $\alpha_{t+1}$ is set adaptively based on the discrepancy between
$S^{\mathrm{pred}}_{t+1}$ and $S^{\mathrm{few}}_{t+1}$ and the few shot sampling
ratio. Let
$\epsilon_{t+1} = \bigl| S^{\mathrm{pred}}_{t+1} - S^{\mathrm{few}}_{t+1} \bigr|$
, and let
$\tau > 0$ be a calibration tolerance. We also define the few shot ratio
$\psi = \frac{|\mathcal{D}_{\mathrm{few}}|}{|\mathcal{D}_{\mathrm{val}}|}$,
and an upper bound $\alpha_{\max}\in(0,1]$ on the calibration strength.
We set $\alpha_{t+1}$ as:
\begin{equation}
\label{eq:alpha_rule}
\alpha_{t+1} =
\begin{cases}
0, & \text{if } \epsilon_{t+1} \le \tau,\\[4pt]
\min\!\Bigl(\dfrac{\epsilon_{t+1}}{\tau}\,\psi,\ \alpha_{\max}\Bigr),
& \text{otherwise.}
\end{cases}
\end{equation}
% Main results
\begin{table*}[htbp]
\centering
\renewcommand{\arraystretch}{1.2}
\small
\setlength{\tabcolsep}{6pt}
\begin{tabular}{
@{}l
*{6}{c@{\hspace{4.5pt}}c@{\hspace{9pt}}}
@{}
}
\hline
& \multicolumn{2}{c}{\textbf{DROP}}
& \multicolumn{2}{c}{\textbf{HotpotQA}}
& \multicolumn{2}{c}{\textbf{GSM8K}}
& \multicolumn{2}{c}{\textbf{MATH}}
& \multicolumn{2}{c}{\textbf{HumanEval}}
& \multicolumn{2}{c}{\textbf{MBPP}} 
\\
\cline{2-13}
\textbf{Method}
& \textbf{Perf}& \textbf{Cost}
& \textbf{Perf}& \textbf{Cost}
& \textbf{Perf}& \textbf{Cost}
& \textbf{Perf}& \textbf{Cost}
& \textbf{Perf}& \textbf{Cost}
& \textbf{Perf}& \textbf{Cost} \\
\hline
$\mathbf{ScoreFlow}$ & 91.48\% & 2.06e7 & 76.85\% & 2.72e7 & 97.79\% & 2.83e7 & 84.35\% & 4.03e7 & 98.91\%  & 3.86e6 &  89.63\%& 3.93e6\\
$\mathbf{AgentPrune}$ & 89.22\% & 6.65e6 & 76.73\% & 2.81e7 & 95.42\% & 1.07e7 & 81.53\% & 2.82e7 & 98.03\% & 1.65e6 & 88.37\% &  1.34e6 \\
\hline
$\mathbf{Aflow}$ & 90.30\% & 1.40e7 & 77.57\%  & 3.11e7 & 96.58\% & 1.11e7 & 82.92\% & 3.06e7 & 99.24\% & 5.26e6 & 89.74\% & 3.92e6\\
\rowcolor{blue!10}\ourMethod  & 88.96\% & 3.27e6 & 77.41\% & 1.43e7 & 95.83\% & 4.04e6 & 81.70\% & 7.04e6 & 98.47\% & 1.09e6 & 90.32\% & 6.83e5\\
\rowcolor{blue!10}$\mathbf{\Delta}$& -1.34\% & $\mathbf{\downarrow 76\%}$ & -0.16\% & $\mathbf{\downarrow 54\%}$ & -0.75\% & $\mathbf{\downarrow 63\%}$ & -1.22\% & $\mathbf{\downarrow 77\%}$ & -0.77\% & $\mathbf{\downarrow79\%}$ & +0.58\% & $\mathbf{\downarrow83\%}$\\
\hline
$\ourMethod\_S^{\mathrm{pred}}$ & 78.61\% & 4.45e6 & 75.40\% & 1.50e7 & 92.51\% & 7.69e6 & 76.33\% & 7.55e6 & 09.16\% & 2.30e5 & 90.62\% & 5.75e5 \\
$\ourMethod\_S^{\mathrm{few}}$  & 86.06\% & 2.85e6 & 73.05\% & 1.55e7 & 94.22\% & 6.28e6 & 78.10\% & 5.49e6 & 96.95\% & 4.13e5 & 91.20\% & 6.17e5 \\
$\ourMethod\_S^{\mathrm{conf}}$ & 00.00\% & 1.78e6 & 00.00\% & 1.04e7 & 0.00\% & 4.43e6 & 57.61\% & 8.48e6 & 97.71\% & 3.15e5 & 88.86\% & 1.24e6 \\
\hline
\end{tabular}
\caption{
    Test performance \textbf{Perf} and token number \textbf{Cost} comparison across six benchmarks.  
    $\mathbf{\Delta}$ reports the performance change and cost reduction of \ourMethod relative to $\mathbf{Aflow}$. 
}
\label{tab:main_results_all}
\vspace{-0.5em}
\end{table*}
Intuitively, when $S^{\mathrm{pred}}_{t+1}$ and $S^{\mathrm{few}}_{t+1}$ agree within the tolerance $\tau$, we keep $\widehat{S}_{t+1}$ equal to the self prediction ($\alpha_{t+1}=0$). When the discrepancy exceeds $\tau$, we increase $\alpha_{t+1}$ proportionally to how many tolerance units $\epsilon_{t+1}$ spans and to the few shot ratio $\psi$, but cap it by $\alpha_{\max}$. This realizes the heuristic that large disagreements are more likely due to prediction error and should be corrected more aggressively toward the few shot estimate, while still respecting the limited size of $\mathcal{D}_{\mathrm{few}}$.
% 如何把估计的经验回传；

To this end, we replace full validation execution with the calibrated surrogate score $\widehat{S}_i$ in Equation~\ref{eq:calibratedscore_final}. The surrogate scores of low-cost stage $\{\widehat{S}_{i}\}_{i>M}$ are used for selection, expansion, and experience update together with the full-execution scores in warm up stage  $\{S^{\mathrm{exec}}_i\}_{i=1}^M$.

Overall, \ourMethod eliminates full validation runs in the main search phase: only few shot execution is performed to calibrate the optimizer's self prediction. As a result, token number scales with $|\mathcal{D}_{\mathrm{few}}|$ instead of $|\mathcal{D}_{\mathrm{val}}|$, cutting the token cost substantially while maintaining the performance.

% Section 4

% Main results
\section{Experiments}
We empirically evaluate our framework on multiple benchmarks, aiming to answer these two questions:
\textbf{Q1:} Can \ourMethod reduce cost while maintaining performance?
\textbf{Q2:} Why does calibrated prediction approximate full execution score well?  

\subsection{Experimental Setup}
\label{sec:exp_setup}
\paragraph{Agent and Optimizer}
In all experiments, we adopt Qwen-Plus~\cite{hui2024qwen_plus} as base models $\{O_{\theta_i}\}_{i=1}^n$ of executor agents and Qwen3-8B~\cite{yang2025qwen3} as the workflow optimizer $\phi$.

\paragraph{Benchmarks}
We evaluate on six benchmarks spanning diverse domains:DROP~\cite{dua2019drop} and HotpotQA~\cite{yang2018hotpotqa} (multi-hop reasoning),GSM8K~\cite{cobbe2021GSM8K} and MATH~\cite{hendrycks2021math} (mathematical reasoning),HumanEval~\cite{chen2021humaneval} and MBPP~\cite{austin2021mbpp} (program synthesis).
The dataset splits follow ~\citet{zhang2024aflow}.

\paragraph{Baselines}
We compare \ourMethod with both task-level methods Aflow~\cite{zhang2024aflow}, AgentPrune~\cite{zhang2024agent_prune} and query-level method ScoreFlow~\cite{wang2025scoreflow}.
We also include internal ablations that vary only in the surrogate score:  
$\ourMethod\_S^{\mathrm{pred}}$ uses uncalibrated self prediction $S^{\mathrm{pred}}$;  
$\ourMethod\_S^{\mathrm{few}}$ uses few shot score $S^{\mathrm{few}}$;  
$\ourMethod\_S^{\mathrm{conf}}$ uses self-confidence $S^{\mathrm{conf}}$, defined as  
$S_{t+1}^{\mathrm{conf}} = \phi(W_i, \widehat{P}^{\mathrm{optimizer}}_{t+1})$,  
where $\widehat{P}^{\mathrm{optimizer}}_{t+1}$ appends \emph{``Output your confidence on the answer''} to the optimizer prompt $P^{\mathrm{optimizer}}_{t+1}$.  
Unlike $S^{\mathrm{pred}}$, $S^{\mathrm{conf}}$ isolates the effect of our dedicated evaluation prompt $P^{\mathrm{eval}}$ by contrasting it with a minimal modification of the generation prompt.

\paragraph{Metrics}
The main metrics reported are: test performance and overall LLM token number incurred excluding test-time execution. 
Each benchmark's performance metric is the same as in \cite{zhang2024aflow}. 
Importantly, the token number is computed differently across methods to reflect their distinct optimization paradigms: 
for \emph{task-level} approaches, it includes all tokens consumed in evaluating candidate workflows on the validation set to select the single best task-level policy; 
for \emph{query-level} method, it includes tokens used in generating data for training the optimizer $\phi$, evaluating query-level workflows during validation, and generating the query-level workflow at test time.

\subsection{Main Results and Ablation Study}
\label{sec:main_results}
Across all six benchmarks, the results in Table~\ref{tab:main_results_all} show that our method substantially reduces token number while maintaining test performance. 
Compared with Aflow, our method \ourMethod yields an average test performance drop of only 0.61\%, while reducing total token number by 54\% to 83\% across all benchmarks. 
This demonstrates that full validation execution is not necessary for discovering high-quality workflows. 
Relative to AgentPrune~\cite{zhang2024agent_prune}, which lowers token cost through structural pruning but still relies on repeated execution-based evaluation, our method achieves further token number reduction while delivering comparable performance, indicating that replacing the evaluation paradigm itself yields greater savings than modifying the search structure alone.

The ablation variants further clarify where these gains come from. $\ourMethod\_S^{\mathrm{pred}}$ drastically reduces cost but exhibits noticeable performance degradation, suggesting that self prediction alone suffers from model bias. $\ourMethod\_S^{\mathrm{few}}$ improves robustness but still requires larger execution budgets. 
$\ourMethod\_S^{\mathrm{conf}}$ performs very bad across tasks, confirming that naive confidence signals are unreliable surrogates for workflow quality. 
In contrast, using calibrated prediction as surrogate evaluation \ourMethod strikes a stable balance between cost and performance by combining model-based prediction with few shot execution signals.

Overall, these results answer \textbf{Q1} affirmatively: \ourMethod maintains the test performance while reducing searching token number by up to 83\%.

\subsection{Comparing Different Surrogate Evaluation Methods}
\label{sec:metric_compare}

To answer \textbf{Q2} , 
For every workflow generated in a full Aflow's run, we log and compare $S^{\mathrm{exec}}$  together four surrogate scores.
Surrogate scores are intended to take place of $S^{\mathrm{exec}}_{i>M}$ to guide selection and expansion, so a good surrogate should satisfy two properties:
First, the value should be close to $S^{\mathrm{exec}}$. If the scales differ, the search will unfairly favor one side of the two-stages.
Second, it should induce a ranking consistent with $S^{\mathrm{exec}}$ to reliably distinguish workflows.

Let $\{x_t\}_{t=1}^T$ denote the sequence of $S^{\mathrm{exec}}$ along the search progress, and $\{y_t\}_{t=1}^T$ the corresponding surrogate scores.
We quantify agreement between $x$ and $y$ with:

\noindent
\textbf{(1) Pearson correlation:} 
$\mathrm{Pearson}(x,y)=
\frac{\sum_{t}(x_t-\bar{x})(y_t-\bar{y})}
{\sqrt{\sum_{t}(x_t-\bar{x})^2}\sqrt{\sum_{t}(y_t-\bar{y})^2}}$
, which measures linear ranking consistency. The larger pearson correlation means the better linear ranking consistency between two metrics. 

\begin{table}[htbp]
\centering
\small
\begin{tabular}{lccc}
\hline
Method & Pearson $\uparrow$ & DiffCos $\uparrow$ & MAE $\downarrow$ \\
\hline
$S^{\mathrm{conf}}$   & -0.0576 & 0.0377  & 0.0802 \\
$S^{\mathrm{pred}}$   & 0.0517  & 0.1275  & 0.0511 \\
$S^{\mathrm{few}}$  & 0.6827 & 0.6192  & 0.2160 \\
\rowcolor{blue!10}$\widehat{S}$  & 0.5217 & 0.5545 & 0.1634 \\
\hline
\end{tabular}
\caption{
Agreement between surrogate evaluation metrics and full execution.
$\widehat{S}$ strikes a good balance between value accuracy and ranking consistency.
}
\label{tab:approx_quality}
\end{table}

\noindent
\textbf{(2) First-order difference cosine similarity:}
$\mathrm{DiffCos}(x,y)=
\frac{\sum_{t=2}^{T}\Delta x_t\,\Delta y_t}
{\sqrt{\sum_{t=2}^{T}\Delta x_t^2}\sqrt{\sum_{t=2}^{T}\Delta y_t^2}}$
 ,where $\Delta x_t = x_t-x_{t-1}$ and $\Delta y_t = y_t-y_{t-1}$.
It captures whether the direction of round-to-round changes is aligned between two workflow's measure metrics. Similar to pearson correlation, the larger DiffCos means the two metrics are better aligned.

\noindent
\textbf{(3) Mean absolute error (MAE):}
$
\mathrm{MAE}(x,y)=\frac{1}{T}\sum_{t=1}^{T} |x_t-y_t|$
which directly measures value-level approximation quality of the surrogate evaluation methods.

Table~\ref{tab:approx_quality} reports these metrics for different surrogates.
$S^{\mathrm{conf}}$ performs poorly on all metrics.
$S^{\mathrm{pred}}$ achieves lowest MAE but almost zero correlation, indicating that it roughly matches the average scale of $S^{\mathrm{exec}}$ yet fails to order different workflows.
$S^{\mathrm{few}}$ shows strong Pearson correlation and DiffCos but suffers from large MAE due to high variance from small sample size.
$\widehat{S}$ combines the strengths of both: it improves correlation over self prediction while reducing MAE compared with few shot execution.

% Overall, the answer to \textbf{Q2} is that the calibrated prediction $\widehat{S}$ serves as the most effective surrogate for  $S^{\mathrm{exec}}$, because it maintains competitive performance while substantially reducing token cost.
Overall, the answer to \textbf{Q2} is that the calibrated prediction $\widehat{S}$ serves as the most effective surrogate for the full-execution score $S^{\mathrm{exec}}$. As a \emph{surrogate method}, it aligns best with $S^{\mathrm{exec}}$ in both value agreement and ranking consistency; as a \emph{evaluation score} for task-level workflow generation, it maintains a competitive test performance while substantially reducing token cost.

% Section 5

\section{Conclusion}
We revisited agentic workflow generation and arrived at two main insights: (1) query-level workflow generation is not always necessary, as a small pool of task-level workflows already achieves strong coverage, and (2) execution-based task-level evaluation is extremely costly while providing limited benefit.
Motivated by these findings, we developed a task-level workflow generation framework \ourMethod to replace costly evaluation with calibrated self prediction. Across multiple benchmarks, it maintains competitive test performance with an average degradation
of just 0.61\% compared to existing approach while reducing overall token number by up to 83\%.
% Short summary

\section{Limitations}
This work still has some limitations. Although our method avoids the main drawbacks of both query-level and task-level workflow generation methods, it remains primarily based on task-level workflow generation and has not yet fully leveraged the generalization capability of task-level approaches together with the fine-grained adaptability of query-level methods. We also do not assess cross-domain generalization in this work.

\bibliographystyle{acl_natbib}
% \bibliography{custom}

\appendix
\section{Appendix}
\subsection{Self Prediction Prompt Template}
\label{app:prompt_template}
The following prompt template is used to guide an LLM-based evaluator in performing \textit{self-prediction}, i.e., estimating the expected accuracy of a candidate workflow over the entire evaluation dataset before actual execution. The prompt is carefully structured to enforce rigorous static analysis while leveraging historical execution feedback for calibration. Key components include: (1) a clear task definition requiring both justification and a calibrated probability score; (2) contextual information about the dataset, few-shot examples, and the workflow’s code structure; (3) explicit descriptions of allowed LLM-based operators to validate correct usage; (4) reference experiences from prior rounds to enable prediction refinement through error reflection; and (5) strict validation rules (e.g., package imports, prompt definitions, operator interfaces) that trigger immediate failure (score = 0.0) if violated. The output format enforces structured reasoning via <reason> and <box> tags to ensure parseable and consistent responses.
\begin{lstlisting}[language=Python, caption={The prompt template for self prediciton.}, label=lst:prediction_prompt]

SELF_PREDICTION_PROMPT = """
You are an expert evaluator of workflows. 
Your task is to predict the probability that a given workflow will correctly execute on the WHOLE DATASET, 
which represents your estimation of its overall accuracy. 
Respond with a brief explanation first, followed by a single floating-point number between 0.0 and 1.0.

Dataset Description:
<dataset>
{dataset_description}
</dataset>

Few-shot samples of the Dataset (jsonl format):
{dataset_few_shots}

Workflow to evaluate (python code):
<workflow>
{workflow}
</workflow>

Prompt used in the workflow (python code):
<prompt>
{prompt}
</prompt>

The workflow is Python code; the key function is __call__(question), which produces the workflow's response. 
The workflow may call LLM-based operators described below:
<operator_description>
{operator_description}
</operator_description>

Reference experiences:
During the warm-up rounds, several workflows have been executed and evaluated. 
Each record includes the round (the iteration number of the workflow), the score (the actual reward obtained after execution), the prediction (the reward you predicted in the previous round), and the python code of the workflow and prompt. (these workflows use the same operators as shown above in the <operator_description></operator_description>)
These experiences are provided to help you calibrate your future predictions by comparing your past predicted rewards with the actual scores, you can adjust your estimation strategy to make your predicted rewards as close as possible to the real execution results.
<experience>
{experience}
</experience>

**General Instructions for evaluation:**
1. Step by step, carefully check for critical errors that could prevent execution:
   - Package check (VERY IMPORTANT): The workflow code imports the required packages (for example: import numpy, asyncio and other commonly used Python packages). If any package is used in the workflow but missing or commented out, output 0.0 directly and do not continue other checks.
   
   - Prompt check (VERY IMPORTANT): The workflow code uses prompts written in Python format. 
   If the workflow uses no prompts then just continue other checks.
   If the workflow uses prompts, 
   For every prompt referenced in the workflow, you must verify that this prompt is properly defined in the prompt.py file (commonly imported as prompt_custom).
   A prompt is considered properly defined only if: It appears in the prompt.py file without being commented out AND the prompt name matches exactly (including capitalization, underscores, and punctuation).
   If any prompt used in the workflow is missing, misspelled, or commented out in prompt.py, you must immediately output 0.0.
   Check ALL prompts used in the workflow following the same rule.
   
   - Operator check (VERY IMPORTANT): The operator is provided in text description format. If the workflow uses an operator, it must be among the operators defined in <operator_description>. If the workflow uses an undefined operator (including mismatched names, incorrect parameters, or improper usage) OR the parameters passed when using an operator do not comply with the interface requirements defined in operator_description, output 0.0 directly and do not continue other checks.
   
   - Workflow check (VERY IMPORTANT): The __call__ funciton must return the output string of the workflow and the token usage only, more or less is totally wrong. The input of the workflow are only the input string, more or less is totally wrong.
   
2. Step by step, Analyze whether the workflow can logically solve the queries in the WHOLE DATASET.Please carefully analyze the function of each operator and whether the position of each operator can smoothly promote the resolution of the problem.

3. Consider potential hallucinations from the operators, for now, we use {backbone_model} as backbone model in operators.

4. Evaluate carefully and comprehensively across all query types in the WHOLE DATASET.

5. Be fair and rational: do not easily assign 0.0 unless there is a severe problem, and avoid scoring 1.0 with overconfidence.

6. There is no need to focus heavily on output details, such as formatting inconsistencies, since extraction is handled simply or by specialized subsequent steps.

Output format:
- Provide a brief explanation of your reasoning in a <reason> tag. 
- Wrap your final probability in a <box> tag **after** the <reason>.  

For example:
<reason>The workflow correctly calls all operators and uses only defined prompts.</reason>
<box>0.85</box>
"""

\end{lstlisting}

\subsection{Workflows Generated By SCALE}
Here we show the best workflows generated by our method SCALE across six datasets.

\paragraph{Workflow and Prompts for DROP}
\begin{lstlisting}[language=Python, caption={The best workflow generated by SCALE for DROP}, label=lst:DROP_workflow]
from typing import Literal
import workspace_SCALE.DROP.workflows_43.template.operator as operator
import workspace_SCALE.DROP.workflows_43.round_6.prompt as prompt_custom
from scripts.async_llm import create_llm_instance


from scripts.evaluator import DatasetType

class Workflow:
    def __init__(
        self,
        name: str,
        llm_config,
        dataset: DatasetType,
    ) -> None:
        self.name = name
        self.dataset = dataset
        self.llm = create_llm_instance(llm_config)
        self.answer_generate = operator.AnswerGenerate(self.llm)
        self.custom = operator.Custom(self.llm)
        self.sc_ensemble = operator.ScEnsemble(self.llm)
        self.verify_output = operator.Custom(self.llm)  # Reusing Custom as VerifyOutput

    async def __call__(self, problem: str):
        """
        Implementation of the workflow
        """
        # Step-by-step generation using AnswerGenerate
        initial_solution = await self.answer_generate(input=problem)
        initial_thought = initial_solution['thought']
        initial_answer = initial_solution['answer']

        # Refine using custom method with clearer instruction and context
        refined_solutions = []
        for _ in range(5):  # Increase number of refinements for better consensus
            refined_sol = await self.custom(
                input=f"{problem}\n\nInitial Thought: {initial_thought}\nInitial Answer: {initial_answer}",
                instruction=prompt_custom.REFINE_WITH_CONTEXT_AND_ACCURATE_PERCENTAGE_CALCULATION_PROMPT
            )
            refined_solutions.append(refined_sol['response'])

        # Use self-consistency ensemble to select the best solution
        ensemble_result = await self.sc_ensemble(solutions=refined_solutions)
        raw_final_response = ensemble_result['response']
        
        # Verification step to ensure final output format compliance
        verified_solution = await self.verify_output(
            input=raw_final_response,
            instruction=prompt_custom.VERIFY_OUTPUT_FORMAT_PROMPT
        )

        return verified_solution['response'], self.llm.get_usage_summary()["total_tokens"]
\end{lstlisting}
\begin{lstlisting}[language=Python, caption={The prompt used in the executor agents of best workflow generated by SCALE for DROP}, label=lst:DROP_prompt]
VERIFY_OUTPUT_FORMAT_PROMPT = """Ensure the response is properly formatted with the answer() wrapper. If the answer is not wrapped in answer(), add it around the final result. For example:
- If the answer is a number: answer(42)
- If the answer is text: answer(Wilson)
- If the answer is a vector: answer(\\begin{pmatrix} 1 \\\\ 2 \\end{pmatrix})
- If the answer is a range: answer(1-10)
- If the answer is a percentage: answer(95%)

The response should contain only the final answer wrapped in answer() with no additional text or explanation."""

VERIFY_MATH_REASONING_PROMPT = """Check the mathematical reasoning in the solution. Verify that:
1. All calculations are correct
2. The logic follows mathematical principles
3. The steps lead to the correct conclusion
4. The final answer is consistent with the reasoning

If any errors are found, correct them and provide the corrected solution. Ensure the final answer is wrapped in answer() with no additional text."""
\end{lstlisting}

\paragraph{Workflow and Prompts for HotpotQA}
\begin{lstlisting}[language=Python, caption={The best workflow generated by SCALE for HotpotQA}, label=lst:Hotpotqa_workflow]
from typing import Literal
import workspace_SCALE.HotpotQA.workflows.template.operator as operator
import workspace_SCALE.HotpotQA.workflows.round_15.prompt as prompt_custom
from scripts.async_llm import create_llm_instance


from scripts.evaluator import DatasetType

class Workflow:
    def __init__(
        self,
        name: str,
        llm_config,
        dataset: DatasetType,
    ) -> None:
        self.name = name
        self.dataset = dataset
        self.llm = create_llm_instance(llm_config)
        self.custom = operator.Custom(self.llm)
        self.answer_generate = operator.AnswerGenerate(self.llm)
        self.sc_ensemble = operator.ScEnsemble(self.llm)
        self.refine = operator.Custom(self.llm)  # New operator for post-ensemble refinement

    async def __call__(self, problem: str):
        """
        Implementation of the workflow
        """
        # Generate multiple reasoning paths
        solutions = []
        for _ in range(3):
            solution = await self.answer_generate(input=problem)
            solutions.append(solution['answer'])

        # Self-consistency ensemble to choose most frequent answer
        ensemble_response = await self.sc_ensemble(solutions=solutions)
        selected_answer = ensemble_response['response']

        # Verify ensemble result for validity and confidence before refining
        verify_response = await self.custom(
            input=f"Question: {problem}\nCandidate Answer: {selected_answer}",
            instruction=prompt_custom.VERIFY_ENSEMBLE_CONFIDENCE_PROMPT
        )
        is_valid = "answer(valid)" in verify_response['response']

        if not is_valid:
            fallback_response = await self.custom(
                input=problem,
                instruction=prompt_custom.FALLBACK_ANSWER_GENERATION_PROMPT
            )
            selected_answer = fallback_response['response']

        # Refine the selected answer with a stricter categorical/entity-focused prompt
        refined_response = await self.refine(
            input=f"Question: {problem}\nSelected Answer: {selected_answer}",
            instruction=prompt_custom.REFINE_TO_ENTITY_OR_CATEGORY_PROMPT
        )
        refined_answer = refined_response['response']

        # Check if refined answer is valid; if not, trigger fallback
        if "answer(None)" in refined_answer or not refined_answer.strip():
            fallback_response = await self.custom(
                input=problem,
                instruction=prompt_custom.FALLBACK_ANSWER_GENERATION_PROMPT
            )
            refined_answer = fallback_response['response']

        # Final formatting verification
        verified_result = await self.custom(
            input=f"Question: {problem}\nCandidate Answer: {refined_answer}",
            instruction=prompt_custom.FINAL_FORMAT_VERIFICATION_PROMPT
        )

        final_output = verified_result['response']
        return final_output, self.llm.get_usage_summary()["total_tokens"]

\end{lstlisting}
\begin{lstlisting}[language=Python, caption={The prompt used in the executor agents of best workflow generated by SCALE for HotpotQA}, label=lst:HotpotQA_prompt]
VERIFY_ENSEMBLE_CONFIDENCE_PROMPT = """You are given a question and a candidate answer derived via ensemble. Determine whether the candidate answer is logically consistent with the question and shows sufficient confidence. If the answer is relevant and confident, respond with answer(valid). Otherwise, respond with answer(invalid). Do not explain or rephrase, just evaluate confidence and relevance."""

REFINE_TO_ENTITY_OR_CATEGORY_PROMPT ="""You are given a question and a candidate answer. Your task is to reformulate the candidate answer into a precise categorical label or named entity that best fits the question. Avoid explanatory or vague language. Return only the most accurate concise form, such as a person's name, a location, a historical event, or a categorical label. Do not add punctuation or quotation marks. Always wrap your final output in answer(...). Examples: answer(Polish independence), answer(William Shakespeare), answer(no), answer(chronological collection of critical quotations)."""

FINAL_FORMAT_VERIFICATION_PROMPT ="""You are tasked with extracting and formatting the final answer from a candidate answer such that it precisely matches the expected format. Avoid including any descriptive or explanatory text. Focus on named entities, binary responses, or categorical labels as appropriate. For named entities (people, places, works), return only the name. For yes/no questions, return exactly "yes" or "no". For categorical responses, return the exact category. Always wrap your final response in answer(...). Examples: answer(Limbo), answer(no), answer(Southern Isles). If the candidate answer contains multiple possibilities, choose the most likely one based on the question. If the candidate is unclear, make a best-guess effort to extract the intended answer. Do not add quotes or extra punctuation. Do not explain your choice."""

FALLBACK_ANSWER_GENERATION_PROMPT ="""Given the original question, please generate a concise and direct answer focusing strictly on the key entity or fact requested. Avoid explanations or additional commentary. Always wrap your final output in answer(...). Example: Question: Who wrote Pride and Prejudice? Output: answer(Jane Austen)"""
\end{lstlisting}

\paragraph{Workflow and Prompts for GSM8K}
\begin{lstlisting}[language=Python, caption={The best workflow generated by SCALE for GSM8K}, label=lst:GSM8K_workflow]
from typing import Literal
import workspace_calibrated_prediction.GSM8K.workflows.template.operator as operator
import workspace_calibrated_prediction.GSM8K.workflows.round_7.prompt as prompt_custom
from scripts.async_llm import create_llm_instance


from scripts.evaluator import DatasetType

class Workflow:
    def __init__(
        self,
        name: str,
        llm_config,
        dataset: DatasetType,
    ) -> None:
        self.name = name
        self.dataset = dataset
        self.llm = create_llm_instance(llm_config)
        self.custom = operator.Custom(self.llm)
        self.programmer = operator.Programmer(self.llm)
        self.sc_ensemble = operator.ScEnsemble(self.llm)

    async def __call__(self, problem: str):
        """
        Implementation of the workflow
        """
        # Step 1: Generate multiple solutions using Programmer for diverse computation paths
        solutions = []
        for _ in range(3):
            solution = await self.programmer(problem=problem, analysis="Solve the following math problem precisely. Return only the final numeric result.")
            solutions.append(solution['output'])

        # Step 2: Use ScEnsemble to select the most consistent solution among the generated ones
        ensemble_result = await self.sc_ensemble(solutions=solutions, problem=problem)

        # Step 3: Format the selected result properly using Custom to ensure it meets expected structure
        formatted_solution = await self.custom(
            input=f"Problem: {problem}\nComputed Result: {ensemble_result['response']}",
            instruction=prompt_custom.FORMAT_ANSWER_PROMPT
        )

        # Step 4: Validate that the result makes sense in context (e.g., not negative where inappropriate, correct order of magnitude)
        validated_solution = await self.custom(
            input=f"Problem: {problem}\nFormatted Result: {formatted_solution['response']}",
            instruction=prompt_custom.VALIDATE_NUMERIC_RESULT_PROMPT
        )

        # Step 5: Final formatting verification to ensure answer is boxed correctly
        final_result = await self.custom(
            input=f"Problem: {problem}\nValidated Result: {validated_solution['response']}",
            instruction=prompt_custom.FINAL_BOXING_CHECK_PROMPT
        )

        return final_result['response'], self.llm.get_usage_summary()["total_tokens"]

\end{lstlisting}
\begin{lstlisting}[language=Python, caption={The prompt used in the executor agents of best workflow generated by SCALE for GSM8K}, label=lst:GSM8K_prompt]
FORMAT_ANSWER_PROMPT ="""You are given a math problem and its computed numeric result. Your task is to format the result in a standardized way by placing it inside \\boxed{}. Only return the final formatted answer without any additional text or explanation. For example, if the result is 123, return \\boxed{123}."""


VALIDATE_NUMERIC_RESULT_PROMPT ="""You are given a math word problem and a formatted numeric result. Check whether the result is logically reasonable in the context of the problem (e.g., not negative when expecting a count, correct magnitude). If it seems incorrect, estimate a plausible value and return it in the same \\boxed{} format. Otherwise, return the original result in \\boxed{} format. Only return the final result in \\boxed{}."""


FINAL_BOXING_CHECK_PROMPT ="""You are given a math problem and a validated numeric result. Ensure that the final result is enclosed in \\boxed{} and represents a clean numeric answer without any extra commentary or formatting issues. Return only the properly boxed result."""
\end{lstlisting}

\paragraph{Workflow and Prompts for MATH}
\begin{lstlisting}[language=Python, caption={The best workflow generated by SCALE for MATH}, label=lst:prompt]
from typing import Literal
import workspace_SCALE.MATH.workflows.template.operator as operator
import workspace_SCALE.MATH.workflows.round_20.prompt as prompt_custom
from scripts.async_llm import create_llm_instance


from scripts.evaluator import DatasetType

class Workflow:
    def __init__(
        self,
        name: str,
        llm_config,
        dataset: DatasetType,
    ) -> None:
        self.name = name
        self.dataset = dataset
        self.llm = create_llm_instance(llm_config)

        self.custom = operator.Custom(self.llm)
        self.verify_format = operator.Custom(self.llm)
        self.verify_math = operator.Custom(self.llm)

    async def __call__(self, problem: str):
        """
        Implementation of the workflow
        """
        solution = await self.custom(input=problem, instruction="")
        
        # Verify mathematical reasoning
        verified_solution = await self.verify_math(input=solution['response'], instruction=prompt_custom.VERIFY_MATH_REASONING_PROMPT)
        
        # Verify and format the output to ensure it has answer() wrapper
        formatted_solution = await self.verify_format(input=verified_solution['response'], instruction=prompt_custom.VERIFY_OUTPUT_FORMAT_PROMPT)
        
        return formatted_solution['response'], self.llm.get_usage_summary()["total_tokens"]

\end{lstlisting}
\begin{lstlisting}[language=Python, caption={The prompt used in the executor agents of best workflow generated by SCALE for MATH}, label=lst:HotpotQA_prompt]
VERIFY_OUTPUT_FORMAT_PROMPT = """Ensure the response is properly formatted with the answer() wrapper. If the answer is not wrapped in answer(), add it around the final result. For example:
- If the answer is a number: answer(42)
- If the answer is text: answer(Wilson)
- If the answer is a vector: answer(\\begin{pmatrix} 1 \\\\ 2 \\end{pmatrix})
- If the answer is a range: answer(1-10)
- If the answer is a percentage: answer(95%)

The response should contain only the final answer wrapped in answer() with no additional text or explanation."""

VERIFY_MATH_REASONING_PROMPT = """Check the mathematical reasoning in the solution. Verify that:
1. All calculations are correct
2. The logic follows mathematical principles
3. The steps lead to the correct conclusion
4. The final answer is consistent with the reasoning

If any errors are found, correct them and provide the corrected solution. Ensure the final answer is wrapped in answer() with no additional text."""
\end{lstlisting}

\paragraph{Workflow and Prompts for HumanEval}
\begin{lstlisting}[language=Python, caption={The best workflow generated by SCALE for HumanEval}, label=lst:HumanEval_workflow]
from typing import Literal
import workspace_SCALE.HumanEval.workflows.template.operator as operator
import workspace_SCALE.HumanEval.workflows.round_11.prompt as prompt_custom
from scripts.async_llm import create_llm_instance


from scripts.evaluator import DatasetType

class Workflow:
    def __init__(
        self,
        name: str,
        llm_config,
        dataset: DatasetType,
    ) -> None:
        self.name = name
        self.dataset = dataset
        self.llm = create_llm_instance(llm_config)
        self.custom = operator.Custom(self.llm)
        self.custom_code_generate = operator.CustomCodeGenerate(self.llm)
        self.sc_ensemble = operator.ScEnsemble(self.llm)
        self.test = operator.Test(self.llm)

    async def __call__(self, problem: str, entry_point: str):
        """
        Implementation of the workflow
        Custom operator to generate anything you want.
        But when you want to get standard code, you should use custom_code_generate operator.
        """
        # Rephrase the problem for clarity
        rephrased_problem = await self.custom(input="", instruction=prompt_custom.REPHRASE_PROBLEM_PROMPT + problem)
        clarified_problem = rephrased_problem['response']

        # Generate multiple solutions for ensemble
        solutions = []
        tested_solutions = []
        for _ in range(5):
            solution = await self.custom_code_generate(problem=clarified_problem, entry_point=entry_point, instruction="")

            # Reflect on the generated solution to improve it
            reflection_prompt = f"Review the following code solution and fix any logical or syntax errors:\\n\\n{solution['response']}"
            reflected_solution = await self.custom(input=clarified_problem, instruction=reflection_prompt)

            solutions.append(reflected_solution['response'])
            
            # Pre-test each refined solution to filter valid ones early
            test_result = await self.test(problem=problem, solution=reflected_solution['response'], entry_point=entry_point)
            if test_result['result']:
                tested_solutions.append(test_result['solution'])
        
        # Prioritize validated solutions; fallback to all if none pass
        if tested_solutions:
            ensemble_input = tested_solutions
        else:
            ensemble_input = solutions
            
        # Use ScEnsemble to select the most consistent solution
        ensemble_result = await self.sc_ensemble(solutions=ensemble_input, problem=problem)
        final_solution = ensemble_result['response']
        
        return final_solution, self.llm.get_usage_summary()["total_tokens"]

\end{lstlisting}
\begin{lstlisting}[language=Python, caption={The prompt used in the executor agents of best workflow generated by SCALE for HumanEval}, label=lst:HumanEval_prompt]
REPHRASE_PROBLEM_PROMPT = """Please rephrase the following programming problem in clearer terms, making sure to highlight the key requirements and expected output format. Problem: """
\end{lstlisting}

\paragraph{Workflow and Prompts for MBPP}
\begin{lstlisting}[language=Python, caption={The best workflow generated by SCALE for MBPP}, label=lst:MBPP_workflow]
from typing import Literal
import workspace_SCALE.MBPP.workflows.template.operator as operator
import workspace_SCALE.MBPP.workflows.round_7.prompt as prompt_custom
from scripts.async_llm import create_llm_instance


from scripts.evaluator import DatasetType

class Workflow:
    def __init__(
        self,
        name: str,
        llm_config,
        dataset: DatasetType,
    ) -> None:
        self.name = name
        self.dataset = dataset
        self.llm = create_llm_instance(llm_config)

        self.custom = operator.Custom(self.llm)
        self.custom_code_generate = operator.CustomCodeGenerate(self.llm)
        self.test = operator.Test(self.llm)

    async def __call__(self, problem: str, entry_point: str):
        """
        Implementation of the workflow
        Custom operator to generate anything you want.
        But when you want to get standard code, you should use custom_code_generate operator.
        """
        # Generate the initial solution
        solution = await self.custom_code_generate(problem=problem, entry_point=entry_point, instruction="Generate Python code that solves the given problem. Make sure to return the result of the function, not just print it.")
        
        # Verify that the solution has proper format and return statements
        verified_solution = await self.custom(input=f"Problem: {problem}\nEntry point: {entry_point}\nSolution:\n{solution['response']}", instruction=prompt_custom.VERIFY_CODE_FORMAT_PROMPT)
        
        # Test the verified solution to ensure it works correctly
        test_result = await self.test(problem=problem, solution=verified_solution['response'], entry_point=entry_point)
        
        final_solution = test_result['solution'] if test_result['result'] else verified_solution['response']
        
        return final_solution, self.llm.get_usage_summary()["total_tokens"]

\end{lstlisting}
\begin{lstlisting}[language=Python, caption={The prompt used in the executor agents of best workflow generated by SCALE for MBPP}, label=lst:MBPP_prompt]
VERIFY_CODE_FORMAT_PROMPT = """Verify that the provided code solution has the correct function signature and includes proper return statements. The solution must:
1. Contain the function with the exact name specified in the entry_point
2. Include a return statement that returns the result of the function (not just print it)
3. Follow proper Python syntax

If the code is missing the function signature or return statement, please fix it. Return the corrected code."""
\end{lstlisting}

\end{document}